\documentclass[submission,copyright,creativecommons]{eptcs}
\providecommand{\event}{ACT 2021} 
\usepackage[export]{adjustbox}
\usepackage{amsfonts}
\usepackage{amsmath}
\usepackage{amsthm}
\usepackage{bm}
\usepackage{booktabs}
\usepackage{enumitem}
\usepackage{graphicx}
\usepackage[htt]{hyphenat}
\usepackage{latexsym}
\usepackage{lmodern}
\usepackage{mathtools}
\usepackage{physics}
\usepackage{stmaryrd}
\usepackage{subfiles}
\usepackage{xcolor}

\newcommand{\lrbracket}[1]{\left\llbracket #1 \right\rrbracket}

\newcommand{\cn}{\text{CN}}
\newcommand{\cnw}{\text{CN}_{\text{word}}}
\newcommand{\cns}{\text{CN}_{\text{string}}}
\newcommand{\ra}[1]{\renewcommand{\arraystretch}{#1}}

\colorlet{color1}{red!90!}
\colorlet{color2}{green!20!orange!100!}

\title{Composing Conversational Negation}
\author{Razin A. Shaikh\thanks{Equal contribution}
\institute{Mathematical Institute,\\
University of Oxford}
\email{razin.shaikh@maths.ox.ac.uk}
\and
Lia Yeh\footnotemark[1]
\institute{Quantum Group,\\
University of Oxford}
\email{lia.yeh@cs.ox.ac.uk}
\and
Benjamin Rodatz\footnotemark[1]
\institute{Computer Science,\\
University of Oxford}
\email{benjamin.rodatz@cs.ox.ac.uk}
\and
Bob Coecke
\institute{Cambridge Quantum}
\email{bob.coecke@cambridgequantum.com}
}

\def\titlerunning{Composing Conversational Negation}
\def\authorrunning{R. A. Shaikh, L. Yeh, B. Rodatz \& B. Coecke}
\begin{document}
\maketitle

\begin{abstract}
Negation in natural language does not follow Boolean logic and is therefore inherently difficult to model.  In particular, a broader understanding of what is being negated must be taken into account.
In previous work, we proposed a framework for the negation of words that accounts for `worldly context'. This paper extends that proposal to now account for the compositional structure inherent in language within the DisCoCirc framework. 
We compose the negations of single words to capture the negation of sentences.  We also describe how to model the negation of words whose meanings evolve in the text.   





\end{abstract}


\section{Introduction}
\label{sec:intro}
Negation in language is a complicated operation.  Differing views of negation in language are a recurring subject of debate amongst linguists, epistemologists, and psychologists.  One view maintains that negation in language conveys denial of a proposition~\cite{Evans1996role}.  Another view on negation in language asserts that it is the collective notion of plausible alternatives~\cite{oaksford:2002contrast}; this view traces its origins to as far back as Plato's view of not-being-X as \emph{otherness}-\mbox{than-X}~\cite{lee:1972plato}.  An explanation compatible with both views is that there can be different stages at which the negation is interpreted, for instance, initially denying information, and later searching for alternatives~\cite{prado:2006reaction}. This search for alternatives differentiates negation in conversation from simple logical negation.

Consider the sentences:
\begin{enumerate}[label=\alph*), topsep=0.1em, itemsep=-0.3em]
    \item \texttt{This is not a hamster; this is a guinea pig.}
    \item \texttt{This is not a hamster; this is a planet.}
\end{enumerate}
Both sentences are grammatically and logically correct.  However, most users of the English language will agree that there is something wrong with sentence b).  Unlike sentence a), which seems reasonable without context, sentence b) must undergo a highly unusual `contextual pressure'~\cite{kruszewski:2016convneg} to be believable---imagine a sci-fi flick about hamster-sized planets. The plausibility of different alternatives to a negated word naturally has a grading~\cite{oaksford:2002contrast, kruszewski:2016convneg}. Our previous work modelled operational conversational negation of a word, and experimentally validated that it positively correlates with human judgment of this grading~\cite{rodatz:2021conversationalnegation}.

This paper will show how to perform the conversational negation of sentences using the conversational negation of words. This is analogous to how Coecke~et~al.~\cite{CSC} developed the compositional categorical framework DisCoCat to obtain the meaning of a sentence from the meaning of its words.
To extend from negation of words to negation of a sentence, a new challenge appears: ambiguity arises not only from the meaning of a negated word but also from which word(s) in the sentence the negation is principally applied to.  As an example, take the sentence ``\texttt{Johnny did not travel to Manchester by train}''~\cite{oaksford:1992reasoning}.  Envision that \texttt{Johnny} is given emphasis---the natural conclusion is that someone else, instead of Johnny, went to Manchester by train.  Correspondingly, if the emphasized word were \texttt{Manchester} or \texttt{train}, then the respective conclusions would be that Johnny went elsewhere or Johnny took another mode of transportation.  Therefore, we note that the conversational negation of this sentence is arrived at from the conversational negation of its constituent words, i.e.~constituent negation \cite{sep-negation}.  We also see that the grammatical structure is unaltered between the non-negated and negated forms of the sentence. This is in line with other attempts in deciphering the meaning of negation in natural language \cite[pg.~104]{kamp:2013discourse}.


Another challenge we tackle in this paper is performing conversational negation of words or entities whose meanings are not fixed; instead, the meanings evolve as we obtain more information in the text. For instance, in the text ``\texttt{Waffle is a dog. Waffle is fluffy. Waffle likes to play fetch.}'', the meaning of the entity ``\texttt{Waffle}'' evolves with each sentence; i.e.~we first learn that ``\texttt{Waffle}'' refers to a dog, then that it is fluffy, and finally, that it likes to play fetch. We want to perform the conversational negation, i.e.~model alternatives, of the entity ``\texttt{Waffle}'' whose meaning updates with time.
The negation of an evolving entity has ambiguity similar to the negation of sentences: we do not know to which information/property of the evolving entity the negation is applied.
Coecke~\cite{coecke:2020textstructure} extended the DisCoCat framework to DisCoCirc, which models evolving meanings by allowing us to compose sentences in the text. We will show that in the DisCoCirc formalism, the negation of evolving meanings is, in fact, the same thing as the negation of sentences. Hence, we will use the same framework of conversational negation that we derived for sentences.

\paragraph{Structure} 
In Section~\ref{sec:background}, we present the compositional DisCoCirc framework and the meaning category of positive operators and completely positive maps. In Section~\ref{sec:cn-word}, we summarize the framework for conversational negation of words that we originally proposed in~\cite{rodatz:2021conversationalnegation}. In Section~\ref{sec:cn-sentence}, we go from conversational negation of words to conversational negation of sentences. In Section~\ref{sec:cn-dynamic}, we model the conversational negation of entities whose meanings evolve in the text. Finally, in Section~\ref{sec:future}, we discuss the implications of the ideas presented in this paper and outline directions for future work.



\section{Compositional language meaning}
\label{sec:background}

\subsection{DisCoCirc}
\label{sec:discocirc}
The DisCoCat framework~\cite{CSC} combines grammar (cf.~categorial grammar~\cite{LambekBook}) and meanings (cf.~vector embeddings in machine learning) within one compositional framework that enables one to compute the meaning of a sentence from the meanings of its words.  To achieve this, it exploits the common compact closed categorical structure, be it of vectors and linear maps, or of positive operators and completely positive maps~\cite{bankova:2019gradedentailment, piedeleu:2015disambiguation}.  The DisCoCirc framework~\cite{coecke:2020textstructure} improved on DisCoCat, (1) by enabling one to compose sentences into larger text, just as gates are composed in circuits; (2) by allowing meanings to evolve as that text evolves; (3) by specifying the sentence type as the tensored spaces of those entities that evolve.  For our purposes, a DisCoCirc diagram has two key ingredients: (1) meaning states; (2) updates~\cite{coecke:2020meaningupdate}:
\begin{equation}
    \includegraphics[scale=0.8, valign=c]{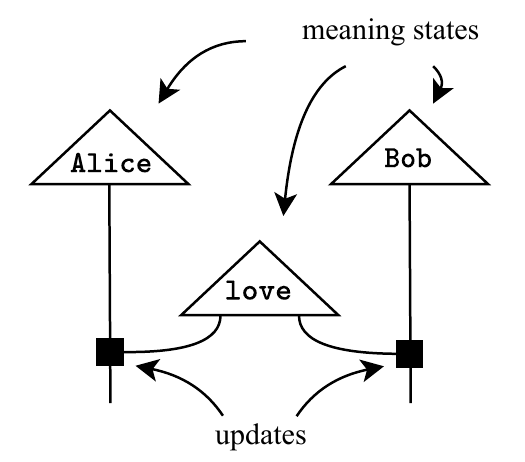} 
    \label{eq:AliceLovesBob1}
\end{equation}
For example, here we have the noun meanings {\tt Alice} and {\tt Bob}, which initially are separate, being updated with the verb meaning {\tt love}.  Alternatively, we can have noun-wires with open input, which we then update to being {\tt Alice} and {\tt Bob} respectively, and then update by {\tt love}:
\begin{equation}
    \qquad\qquad\ \includegraphics[scale=0.8, valign=c]{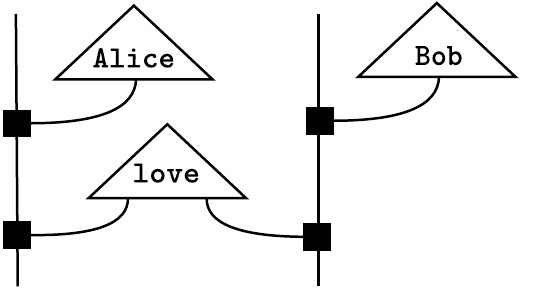} 
    \label{eq:AliceLovesBob2}
\end{equation}
These open input wires allow for pre-composition with other sentences.

\subsection{The meaning category CPM(\textbf{FHilb})}
\label{sec:entailment}

While the DisCoCirc framework allows for various encodings of meaning, for this paper, we work with the category CPM(\textbf{FHilb}) of positive operators and completely positive maps, as done in~\cite{lewis:2020towardslogicalnegation, coecke:2020meaningupdate, piedeleu:2015disambiguation}. Positive operators are complex matrices, which are equal to their own conjugate transpose (Hermitian) and have non-negative eigenvalues (positive semidefinite). Completely positive maps are linear maps from positive operators to positive operators. The compact closed category CPM(\textbf{FHilb}) can be seen as an extension of the category of finite-dimensional Hilbert spaces \textbf{FHilb}. It can be obtained from \textbf{FHilb} via the CPM construction originally introduced by Selinger~\cite{sellinger:2007CPM}. In this construction, for any given unit vector $\ket{v}$ of a finite-dimensional Hilbert space, we can obtain a pure state positive operator by taking the outer product $\ket{v}\bra{v}$. All other positive operators can be obtained as a linear combination of pure states.


In contrast to vectors, which have no inherent ordering structure~\cite{balkir:2016entailment-using-density-matrices}, positive operators can be viewed as an extension of vector spaces that allows for encoding lexical entailment structure such as proposed in~\cite{bankova:2019gradedentailment, lewis:2019compositionalhyponymy}. We use these entailment measures to capture hyponomy: a word $w_1$ is a hyponym of $w_2$ if $w_1$ is a type of $w_2$; then, $w_2$ is a hypernym of $w_1$. For example, \emph{dog} is a hyponym of \emph{animal}, and \emph{animal} is a hypernym of \emph{dog}. These entailment measures are often graded and take values between 0 (no entailment) and 1 (full entailment).

Additionally, positive operators can be used to encode ambiguity---words having multiple meanings---via mixing~\cite{piedeleu:2015disambiguation, meyer-lewis-2020-modelling}, i.e.~taking weighted sums over the different meanings. This ambiguity can later be disambiguated through additional meaning updates \cite{coecke:2020meaningupdate, meyer-lewis-2020-modelling}. We will use this property of positive operators to encode the ambiguity of negation in Section~\ref{sec:cn-sentence}.
For analysis and model performance for encoding and resolving ambiguity in compositional distributional semantics, we refer the reader to the extensive literature~\cite{blacoe-etal-2013-quantum, kartsaklis-etal-2013-separating, kartsaklis-sadrzadeh-2013-prior, piedeleu:2015disambiguation, coecke:2020meaningupdate, meyer-lewis-2020-modelling}.
\section{Conversational negation of words}\label{sec:cn-word}
This section summarizes the operation for conversational negation of words as we proposed in~\cite{rodatz:2021conversationalnegation}. As pointed out in Section~\ref{sec:intro} with the example of hamster-sized planets, negation in conversation not only denies information---a planet is indeed not a hamster---but additionally utilizes the listener's understanding of the world to weigh possible alternatives. This builds on the assumption that plausible alternatives to a word should appear in a similar context~\cite{oaksford:1992reasoning}. To present the framework for conversational negation, we will first present logical negation operations and an encoding of worldly knowledge.

\subsection{Logical negation}
Logical negation (denoted by $\neg$) should fulfill certain properties such as the double negative ($\neg (\neg p) = p$) and the contrapositive ($p \sqsubseteq q \Leftrightarrow \neg q \sqsubseteq \neg p$). 
Lewis~\cite{lewis:2020towardslogicalnegation} proposes the operation $\neg \textsf{X} \coloneqq \mathbb{I} - \textsf{X}$, mapping in the case of projectors to the orthogonal subspace, as Widdows and Peters did for vectors~\cite{widdows:2003word}. In~\cite{rodatz:2021conversationalnegation}, we proposed another logical negation based on generalizing the matrix inverse, that satisfies the contrapositive condition for the (graded) L\"owner order~\cite{bankova:2019gradedentailment}.

In some sense, the result of the logical negation of a word is akin to a mixture of everything that is not that word. Despite this aligning with the set-theoretic notion of complement sets, this is unlike how humans perceive negation.  Indeed, in our prior experiments, we found that alternatives elicited by both of the proposed logical negations have a negative correlation with human intuition~\cite{rodatz:2021conversationalnegation}. We remedied this by amending logical negation with the worldly context of the negated word to achieve a positive correlation.


\subsection{Worldly context}
\label{sec:worldlyContext}

Worldly context is another primary ingredient of conversational negation. It encodes the intuitive understanding of the world most readers possess, by capturing the context in which a word tends to appear. Thus, as it encodes the space of possible alternatives to a word, worldly context can be utilized to weigh the results of the logical negation.

\begin{figure}[htp]
    \centering
    \includegraphics[scale=0.9]{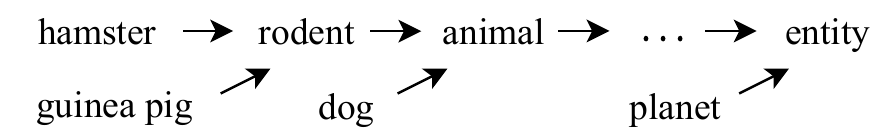}
    \caption{Example of hyponymy structure as can be found in entailment hierarchies}
    \label{fig:tree-hierarchy}
\end{figure}

To build worldly context for a given word, we proposed utilizing entailment hierarchies such as the human-curated WordNet~\cite{wordnet} or the unsupervised Hearst patterns~\cite{hearst:1992automatic}. These hierarchies give us entailment relations such as displayed in Figure~\ref{fig:tree-hierarchy}, 
where each directed edge represents an entailment. We thus get relations such as every hamster being a rodent, every rodent being an animal, and all animals being entities. When negating hamster, we are most likely to talk about other rodents such as guinea pigs. We are less likely to talk about other animals such as dogs and yet less likely to talk about other entities such as planets.

Building on this idea, we proposed to construct the worldly context of a word by considering its hypernym paths and taking a weighted sum over all hypernyms. Hence, for a word $w$ with hypernyms $h_1, \ldots, h_n$ ordered from closest to furthest, we define the worldly context $\texttt{wc}_{w}$ as:
\begin{equation} \label{eq:wc-wordnet}
    \lrbracket{\texttt{wc}_{w}} \coloneqq \sum_i p_i \lrbracket{h_i}
\end{equation}
where $p_i \geq p_{i+1}$ for all $i$.  We denote the positive operator encoding the meaning of a word with double brackets.

\subsection{Framework for conversational negation of words}
\label{sec:frameworkCNword}

In~\cite{rodatz:2021conversationalnegation}, conversational negation of words, written as the operation $\cnw$, is defined as:
\begin{equation}
    \includegraphics[scale=0.6, valign=c]{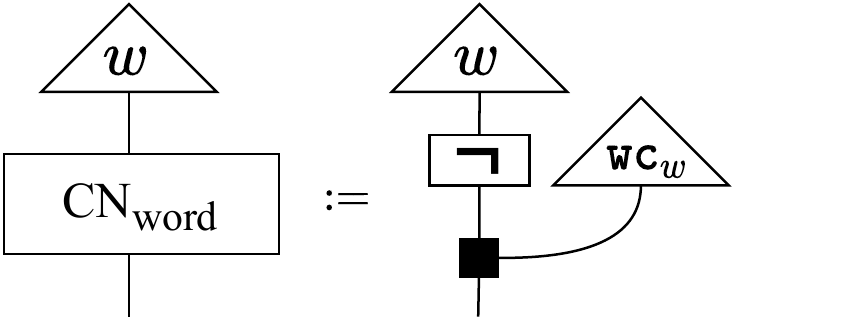} 
    \label{eq:convNegFramework}
\end{equation}
This framework can be interpreted as the following three steps.
\begin{enumerate}[itemsep=0em, topsep=0.2em]
    \item Calculate the logical negation $\neg(\lrbracket{w})$.
    \item Compute the worldly context $\lrbracket{\texttt{wc}_w}$.
    \item Update the meaning of $\neg(\lrbracket{w})$ by composing with $\lrbracket{\texttt{wc}_w}$ to obtain $\neg(\lrbracket{w})\  \includegraphics[height=0.9em]{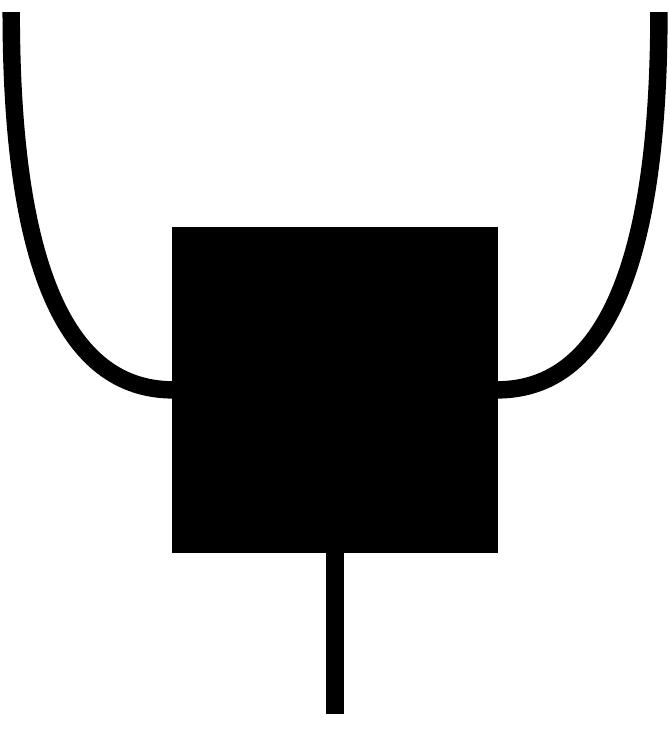} \ \lrbracket{\texttt{wc}_w}$.
\end{enumerate}
This framework is flexible to the choice of logical negation, worldly context generation and composition operation \includegraphics[height=0.9em]{figures/conjunction-symmetric.pdf}. In~\cite{rodatz:2021conversationalnegation}, we studied and compared the performance of various choices of operations. While we only experimentally validated our negation operation on nouns, the same operation is applicable to adjectives and verbs.




\section{Conversational negation of strings of words}\label{sec:cn-sentence}
In the distributional approach to natural language processing, a commonly used model is the bag-of-words that disregards any grammar and treats words as a structureless bag. Coecke et~al.~\cite{CSC} proposed the DisCoCat model, which combines grammar and distributional word meanings in a categorical framework, allowing us to compose the meaning of words to get the meaning of a sentence. In this section, we do the same for conversational negation by introducing the conversational negation of sentences that is calculated using the conversational negation of words.

\subsection{Modelling conversational negation}
As pointed out by Oaksford and Stenning~\cite{oaksford:1992reasoning}, the negation of more complex structures consisting of multiple words may be interpreted as the negation of a subset of the constituents. For example, a sentence such as ``\texttt{Bob did not drive to Oxford by car}'' could be interpreted as:

\begin{enumerate}[label=\alph*), topsep=0.1em, itemsep=-0.3em]
    \item {\ttfamily \underline{Bob} did not drive to Oxford by car\quad - \quad Alice did}
    \item {\ttfamily Bob did not \underline{drive} to Oxford by car\quad - \quad He carpooled}
    \item {\ttfamily Bob did not drive to \underline{Oxford} by car\quad - \quad He drove to London}
    \item {\ttfamily Bob did not drive to Oxford by \underline{car}\quad - \quad He drove a van}
    \item {\ttfamily \underline{Bob} did not \underline{drive} to Oxford by car\quad - \quad Alice carpooled to Oxford}
\end{enumerate}
where the underlines indicate which words are being negated. The last example is one of many possible cases that negate multiple constituents. While some of these alternatives might immediately seem more plausible to the reader, the correct choice is inherently dependent on the context.

Based on this interpretation, that a negation of multiple words is negating a subset of the constituents, we extend our conversational negation framework to a string of words by utilizing conversational negation of individual words (see Section~\ref{sec:cn-word}). As the correct interpretation of which words to negate may not usually be obvious, we create a mixture of all possible interpretations. Therefore, the negation of a string of $n$ words $w_1 \otimes w_2 \otimes ... \otimes w_n$ is a weighted mixture of all the interpretations where only one word is negated, all the interpretations where two words are negated, and so on. We have:
\begin{align}
    \includegraphics[height=6.6em, valign=c]{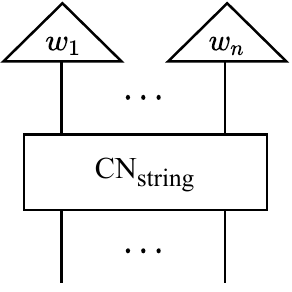} \coloneqq & \quad
    \begin{aligned}
        &\sum_{i = 1}^n p_{\{w_i\}} \left(w_1 \otimes \dots \otimes \cnw(w_i) \otimes \dots \otimes w_n \right) \ +  \\
        & \sum_{i = 1}^n \  \sum_{j = i + 1}^n \,  p_{\{w_i, w_j\}} \left( w_1 \otimes \dots \otimes \cnw(w_i) \otimes \dots \otimes \cnw(w_j) \otimes \dots \otimes w_n \right) \\
        &  + \quad \dots
    \end{aligned}
\end{align}
where in the overall mixture representing the negation, each interpretation has some weight. Formally, for $S = \{w_1, ..., w_n\}$ and non-empty $S' \subseteq S$ which we call the \textbf{negation set}, we get:
\begin{align}
    \includegraphics[height=6.6em, valign=c]{figures/CNstring.pdf} & \coloneqq \quad
    \sum_{S' \in \mathcal{P}(S)\setminus \{\emptyset\}} p_{\scriptscriptstyle S'} \bigotimes_{i=1}^n
    \begin{cases}
        w_i & \text{if } w_i \not \in S'\\
        \cnw(w_i) & \text{if } w_i \in S'
    \end{cases}
\end{align}

To apply this negation to sentences, we adopt the view that sentences are processes updating wires as presented by Coecke in~\cite{coecke:2020textstructure}. These processes are built from a combination of meaning states, interacting via updates. 
We propose the negation of a sentence to be viewed as the same set of meaning states, first updated by the conversational negation of the words before updating the wires as if the negation was absent. This is in line with Kamp and Reyle~\cite[pg.~104]{kamp:2013discourse}, who also treat the grammatical structure of negated sentences as if they were not negated.
Hence, we represent the conversational negation as a function called $\cn$ that maps the circuit of a sentence to a circuit where the sentence is pre-composed with $\cns$.

For example, applying $\cn$ to the sentence ``\texttt{Alice loves Bob}'',  we get the circuit for the sentence ``\texttt{Alice doesn't love Bob}'':
\begin{equation*}
    \includegraphics[width=0.9\textwidth]{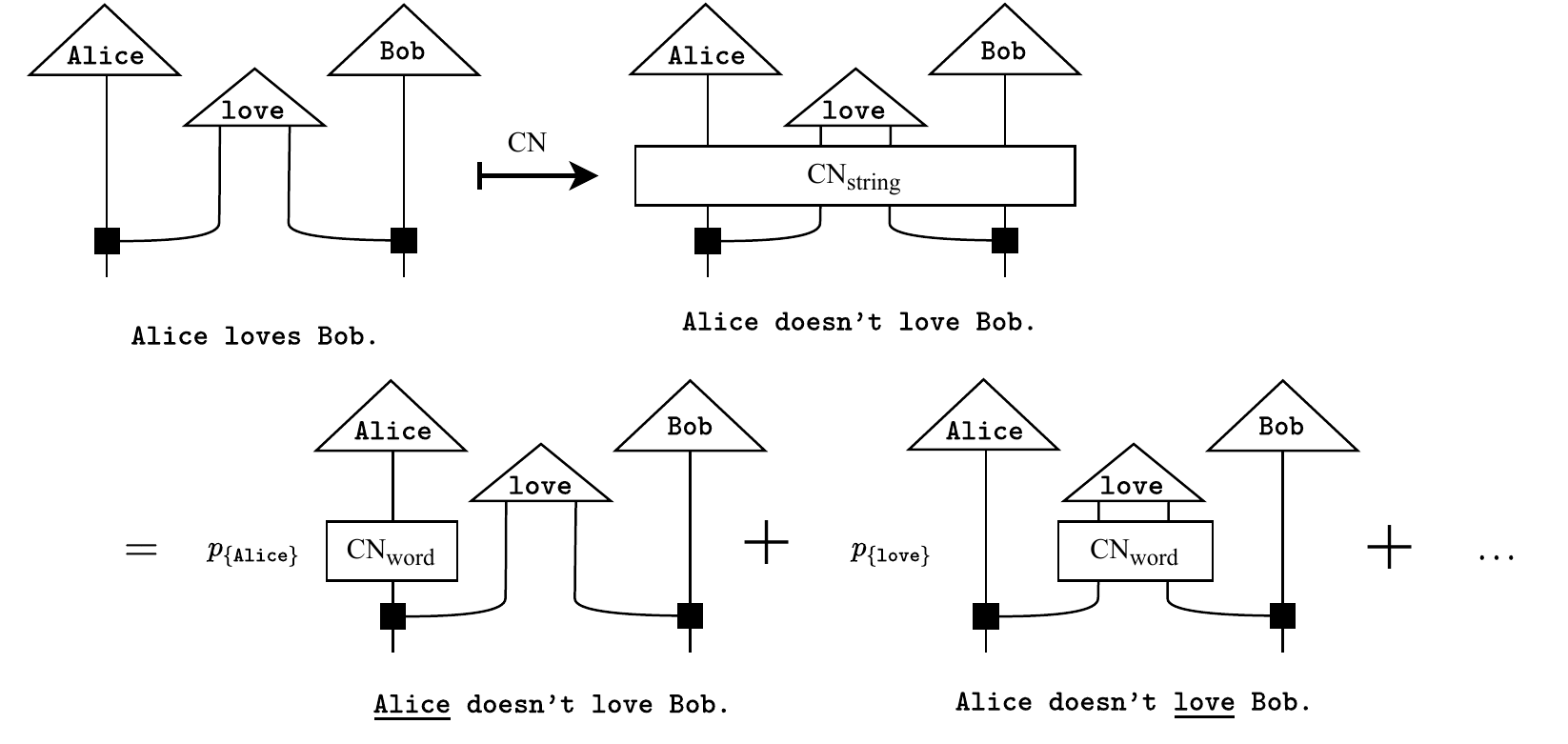}
    \label{eq:aliceDoesntLoveBob}
\end{equation*}
Here, we sum over all possible non-empty subsets of \{\texttt{Alice}, \texttt{love}, \texttt{Bob}\}, each of which is weighted by the appropriate scalar.

\subsection{Deriving the weights}\label{sec:weights}
The main challenge for the conversational negation of a string is deriving the weights for the different interpretations of the negation. The choice of correct interpretation, and therefore the weights, is dependent on context. Context can be derived from many sources, such as the person who is speaking and their intentions. In spoken language, intonation could clarify the speaker's intent by emphasizing the words meant to be negated.

Another source of context is the grammatical structure of the negated sentence itself. Given the earlier example ``\texttt{Bob did not drive to Oxford by car},'' which mentions the mode of transport explicitly, intuitively the focus of negation is on this detail. If the speaker solely wanted to negate Bob's destination, the sentence ``\texttt{Bob did not drive to Oxford}'' would be sufficient, not requiring any additional detail. The other example, ``\texttt{Alice does not love Bob},'' is more ambiguous as the grammatical structure does not indicate the target of the negation.

Context from surrounding sentences can further determine the sensibility of interpretations to a negation. Consecutive sentences should have meanings consistent with each other and the reader's general understanding of the world. The statement \texttt{Bob did not drive to \underline{Oxford} by car - He drove to London} is reasonable with the knowledge that \texttt{Oxford} and \texttt{London} are cities an hour's drive apart.  Likewise, \texttt{Bob did not drive to Oxford by \underline{car} - He drove a van} makes sense because \texttt{car} and \texttt{van} are similar vehicles.

Overall, no single source of context is sufficient. A combination of all contextual information---worldly, textual, physical, grammatical, intonation, etc.---is required to determine the correct interpretation.

While the different weights for the negation sets are context-dependent, some general observations can be made.  Larger negation sets should tend to have smaller weights. The psychological reasoning is that humans are less able to focus on a larger number of details due to ``limited information processing capacity''~\cite{evans1989bias, oaksford:1992reasoning}. Considering the previous example, one would require a lot of context for the interpretation of ``\texttt{{Alice} doesn't {love} {Bob}}'' to sensibly imply ``\texttt{Claire likes Dave}''. Secondly, one can observe that the weight of a negation set should depend on the likelihood of its individual elements to be the target of the negation. If \texttt{Alice} being the target of the negation is unlikely, for instance if the entire text is about \texttt{Alice}, then the negation set of both \texttt{Alice} and \texttt{love} is also unlikely.

\subsubsection{Determining weights using entailment}
\label{section:probabilitiesUsingEntailment}
As mentioned earlier, one possible source of context can be the surrounding text. 
In a text which solely talks about \texttt{Alice} and \texttt{Bob}, the sentence ``\texttt{Alice doesn't love Bob}'' probably intends to negate the word \texttt{love}, therefore asserting that \texttt{Alice} feels emotions other than love for \texttt{Bob}.  Building on this intuition, we propose to use entailment measures to derive the weights for the different interpretations of a negation. If the given interpretation of the negation entails the surrounding sentences to a high degree, then the interpretation is consistent with the surrounding text, and hence it is more likely to be the intended meaning of the sentence. 

We compare each possible interpretation of the negation with the surrounding sentences, where sentences closer in the text have more influence towards the final weighting than sentences that are further away. 
Let us consider the following, simplified scenario of a negation, followed immediately by the clarification with both sentences of the same grammatical structure:
\begin{center}
    {\color{color1} \texttt{This is not red wine}}\\
    {\color{color2} \texttt{This is white wine}}
\end{center}
Here, we colour code the sentences for visual clarity, where the negated sentence is red.
We want to determine the respective weights $p_{\{{\color{color1}\texttt{red}}\}},\ p_{\{{\color{color1}\texttt{wine}}\}}$ and $p_{\{{\color{color1}\texttt{red},\ \texttt{wine}}\}}$ of the negation sets \{{\color{color1}\texttt{red}}\}, \{{\color{color1}\texttt{wine}}\} and \{{\color{color1}\texttt{red}, \texttt{wine}}\}.\footnote{For the sake of simplicity, we ignore the fact that ``This" could also be the target of the negation, as in ``This is not red wine. That is!".}
As a human reader, the sentence following the negation clarifies that \texttt{{\color{color1}wine}} is not negated. Thus, the intended negation set is \{{\color{color1}\texttt{red}}\}.

To mathematically come to the same conclusion, we calculate the entailment between each interpretation of the negation and the follow-up sentence, i.e.~how much each negation set of {\color{color1} {\texttt{not red wine}}} entails {\color{color2} \texttt{white wine}}. We compare the two sentences word by word (or more precisely, treating adjectives and nouns individually) and then take the product of the results. We consider:

\begin{itemize}
    \item {\color{color1} \texttt{not \underline{red} wine}} $\sqsubseteq$ {\color{color2} \texttt{white wine}} - 
    We first calculate the entailment of $\cnw(\texttt{\color{color1} red})$ with $\texttt{\color{color2} white}$ which is medium as something that is not red could have many other colors, including white. The entailment of $\texttt{{\color{color1} wine}}$ with $\texttt{{\color{color2} wine}}$ is maximal as a wine is indeed a wine. Therefore the overall score of this interpretation is high. \\
    \emph{Overall entailment: high}
    \item {\color{color1} \texttt{not red \underline{wine}}} $\sqsubseteq$ {\color{color2} \texttt{white wine}}  - This interpretation has medium entailment between \\$\cnw(\texttt{\color{color1} wine})$ and $\texttt{\color{color2} wine}$ due to the fact that in distributional semantics, a word and its conversational negation appear in similar contexts \cite{saif:2013computingLexicalContrast, oaksford:1992reasoning}. Yet the entailment between $\texttt{{\color{color1} red}}$ and $\texttt{{\color{color2} white}}$ is low since something being red does not entail that it is white. \\
    \emph{Overall entailment: low}
    
    \item {\color{color1} \texttt{not \underline{red} \underline{wine}}} $\sqsubseteq$ {\color{color2} \texttt{white wine}} - This interpretation has a medium entailment between $\cnw(\texttt{\color{color1} red})$ and $\texttt{\color{color2} white}$ and a medium entailment between  $\cnw(\texttt{\color{color1} wine})$ and $\texttt{\color{color2} wine}$.\\
    \emph{Overall entailment: medium}
\end{itemize}
Comparing the three interpretations, the first option has the highest score, matching our intuition of being the correct choice. 

While this entailment method relies on the sentences having an identical grammatical structure to compare the sentences word by word, we can also directly compare entailment between two sentences. This requires composition operations and entailment measures which interact well to preserve word level entailment on the sentence level. This is still a field of active research, with some promising results presented in \cite{kartsaklis-sadrzadeh-2016-distributional, kartsaklis-etal-2016-distributional, Sadrzadeh_2018, delascuevas:2020catsclimb}.


\section{Conversational negation of evolving meanings}\label{sec:cn-dynamic}

Coecke~\cite{coecke:2020textstructure} enhanced the DisCoCat framework to create DisCoCirc, which allows the meaning of sentences to be composed to obtain the meaning of text. In this section, we show that our conversational negation framework can be easily extended from sentences to conversational negation of text and evolving meanings.

\subsection{Negating evolving meanings}
\label{sec:cn-dynamic-framework}


One of the key features of DisCoCirc is that it allows the meanings of entities to evolve as text evolves. The meanings are updated when the wires of the entities are composed with some meaning states. In DisCoCirc, texts that have the same meaning result in the same updates on the wires, even if they contain different sentences. For example, a text containing the following two sentences:
\begin{center}
    \texttt{Bob is a scientist.}\\
    \texttt{Bob is an alcoholic.}
\end{center}
results in the same circuit as the text containing the single sentence:
\begin{center}
    \texttt{Bob who is a scientist is an alcoholic.}
\end{center}
This motivates us to think of all the meaning updates to a wire as a single large sentence. If we have a sequence of updates to a wire, we can morph the wire using the snake equation. We thereby derive a single meaning update process that updates the wire with a sequence of word meanings. Hence, for a wire whose meaning evolves through updates by words $w_1, \cdots, w_n$, we have:
\begin{equation}
    \includegraphics[scale=0.85, valign=c]{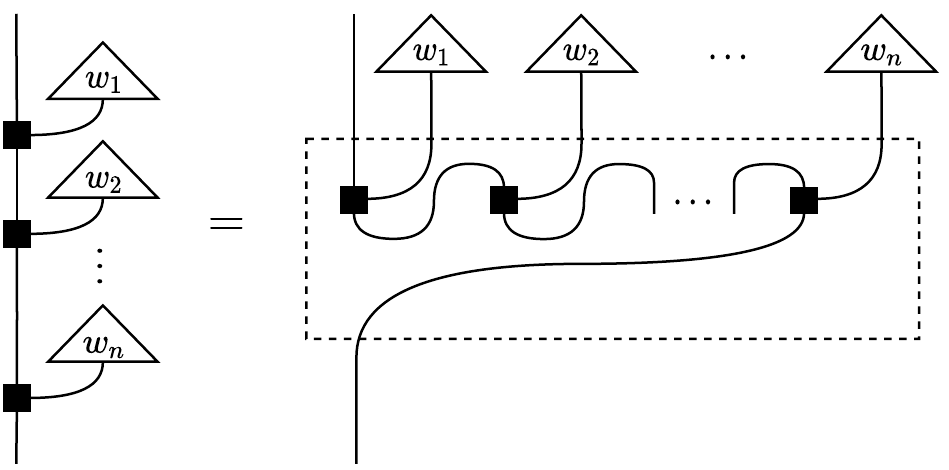}
\end{equation}
Now, to perform conversational negation, we can simply apply the $\cn$ function (see Section~\ref{sec:cn-sentence}) that maps the circuit of a sentence to a circuit where the sentence is pre-composed with $\cns$. Thus, the conversational negation of a dynamic evolving entity becomes:
\begin{equation}
    \includegraphics[scale=0.85, valign=c]{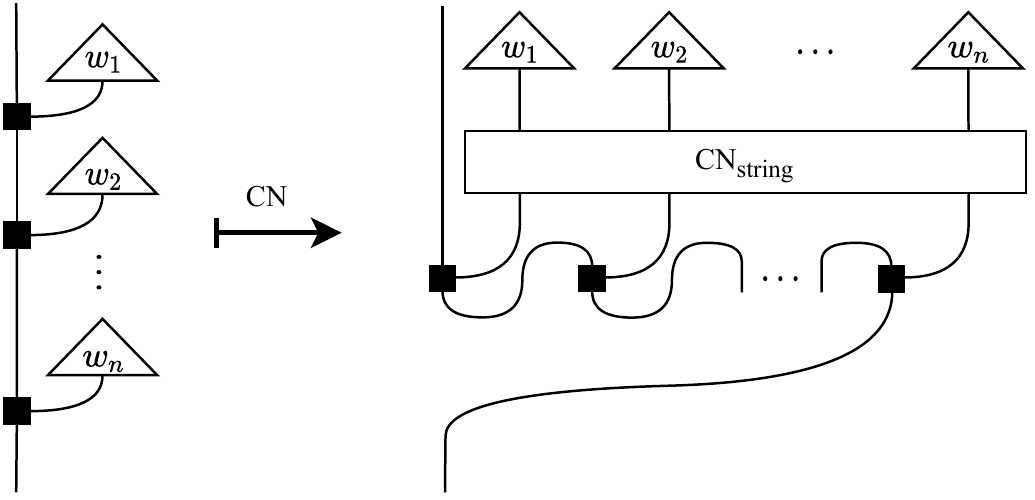}
    \label{eq:cn-wire-fig}
\end{equation}

\noindent
This idea can also be applied to text with multiple, interacting actors. Consider the text:
\begin{center}
    \texttt{Alice is evil.}\\
    \texttt{Bob is old.}\\
    \texttt{Alice loves Bob.}
\end{center}
Alternatively, we can write this as a single sentence, omitting commas for simplicity:
\begin{center}
    \texttt{Alice who is evil loves Bob who is old.}
\end{center}
The circuit for the text looks like the following:
\begin{equation*}
    \includegraphics[scale=0.75, valign=c]{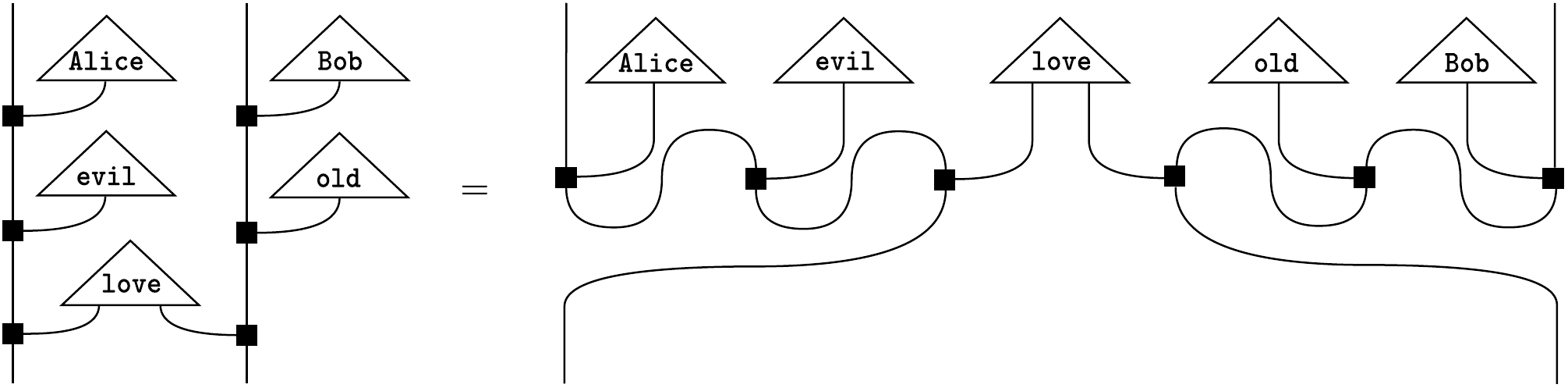}
\end{equation*}
Suppose we want to find the conversational negation of the actor named Alice in the above text, i.e.~we want to know that if someone is not Alice, who else could they be. Similar to the case of conversational negation of sentences, we have various possibilities based on the meaning updates that have been performed on the wire corresponding to Alice. Any subset of the words that have contributed to the meaning update of Alice --- either directly or via meaning updates to entangled actors --- could be negated. For example, possibilities of who ``not \texttt{Alice}'' could be include:
\begin{enumerate}[label=\alph*), topsep=0.1em, itemsep=-0.3em]
    \item {\ttfamily \underline{Claire} who is evil and loves old Bob}
    \item {\ttfamily Alice (different person but same name) who is \underline{virtuous} and \underline{hates} old Bob}
    \item {\ttfamily \underline{Dave} who is evil and loves \underline{young} Bob}
    \item {\ttfamily \underline{Claire} who is \underline{virtuous} and \underline{hates} old \underline{Daisy}}
\end{enumerate}
If we again think of the text as a single large sentence, we can simply apply the $\cn$ function from Section~\ref{sec:cn-sentence}. We get:
\begin{equation*}
    \includegraphics[scale=0.75, valign=c]{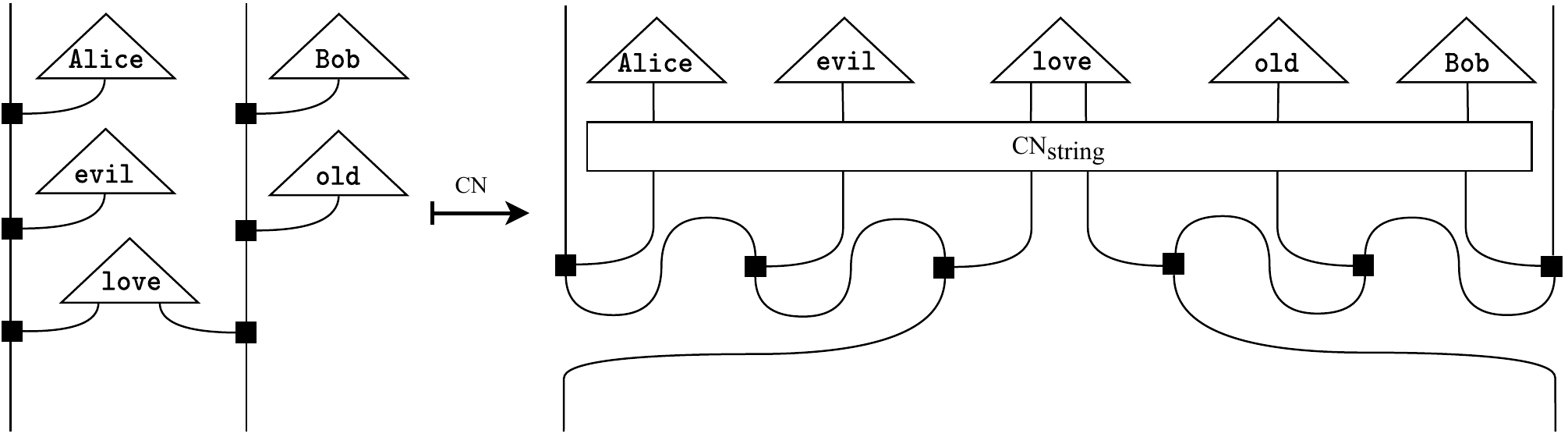}
\end{equation*}
\begin{equation*}
    \includegraphics[scale=0.75, valign=c]{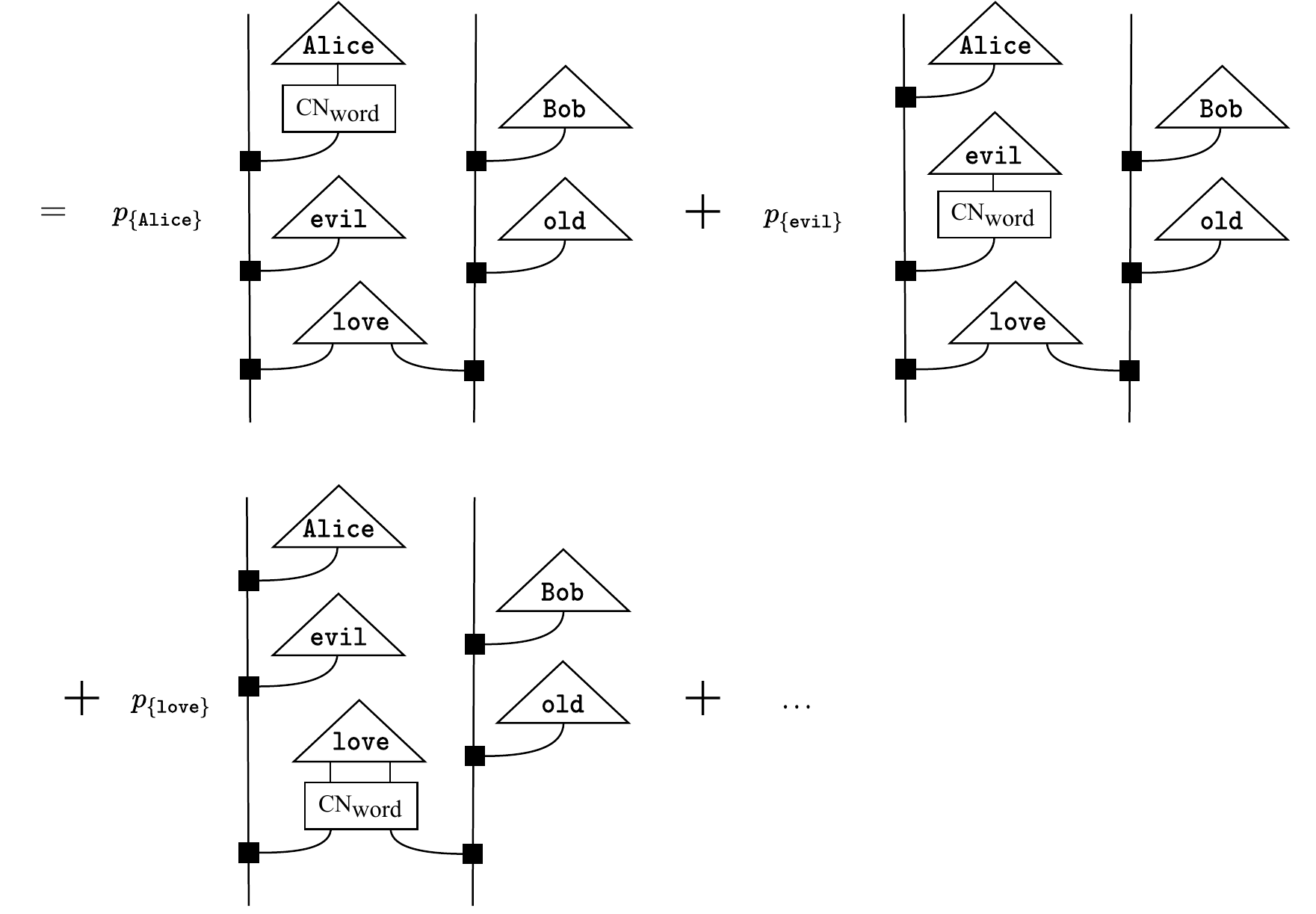}
\end{equation*}
If we want to perform the conversational negation of the actor \texttt{Bob} instead of \texttt{Alice}, we will again get the same resulting circuit after applying the $\cn$ function. The difference to the conversational negation of \texttt{Alice} is that we will have different weights for the negation sets. For example, the weight of the negation set \{\texttt{Bob}, \texttt{old}\} is likely low, for the negation of \texttt{Alice} --- someone who is not \texttt{Alice} --- is intuitively unlikely to be someone who is called Alice and is evil but loves a person with different attributes than \texttt{Bob}. In contrast, the weight of the same negation set could be reasonably high, for the negation of \texttt{Bob} --- someone who is not Bob --- could have a different name and age but still be the lover of \texttt{Alice}. Therefore, while having the same diagrammatic representation, the negation of the actor \texttt{Alice} can give a fundamentally different result than the negation of \texttt{Bob} through the weights which were chosen for the negation sets.

\subsection{Example: finding alternatives}
Consider the following text where the meanings of the words evolves as the text evolves:
\begin{align*}
    \texttt{Alice is a human.}
    \qquad&\texttt{Alice is an archaeologist.}\\
    \texttt{Bob is a human.}
    \qquad&\texttt{Bob is a biologist.}\\
    \texttt{Claire is a human.}
    \qquad&\texttt{Claire is a pianist.}\\
    \texttt{Daisy is a dog.}
    \qquad&\texttt{Daisy is a pet.}
\end{align*}
Suppose we want to perform the conversational negation of the actor \texttt{Alice} and evaluate how much it entails the other actors: \texttt{Bob}, \texttt{Claire} and \texttt{Daisy}. Based on the given text, it is reasonable to expect that someone who is not \texttt{Alice} (a human archaeologist) is more likely to be \texttt{Bob} (a human biologist) than \texttt{Claire} (a human pianist). In fact, someone who is not \texttt{Alice} is still more likely to be \texttt{Claire} (a human pianist) than \texttt{Daisy} (a pet dog). Now we will analyze if the conversational negation presented in Section~\ref{sec:cn-dynamic-framework} reflects this intuition.

When we apply the conversational negation to the actor \texttt{Alice}, we get a mixture containing all possible negation sets along with their weights. These negation sets are nonempty subsets of \{\texttt{Alice}, \texttt{human}, \texttt{archaeologist}\}. The weights of the negation sets must be determined using all contexts as discussed in Section~\ref{sec:weights}.
However, to explore the maximum entailment that can be achieved from the negation of the actor \texttt{Alice} to each of the remaining actors, we only consider the most appropriate negation sets of ``not \texttt{Alice}'' for each actor.

\begin{itemize}[leftmargin=2em]
    \item \texttt{Bob} - Since \texttt{Bob} is a human biologist, the best negation set of ``not \texttt{Alice}'' for \texttt{Bob} is \{\texttt{Alice}, \texttt{archaeologist}\}.  The table below shows the entailment between this negation set of ``not \texttt{Alice}'' and the actor \texttt{Bob}. From the table, it is clear that ``not \texttt{Alice}'' \textbf{highly} entails \texttt{Bob}.
    
    \begin{table}[!ht]
        \ra{1.2}
        \setlength{\tabcolsep}{18pt}
        \centering
        \begin{tabular}{@{}lll@{}}
            \toprule
            ``not \texttt{Alice}''         & \texttt{Bob}        & Entailment  \\ \hline
            $\cnw(\texttt{Alice})$         & \texttt{Bob}        & medium      \\ 
            \texttt{human}                 & \texttt{human}      & 1 (max)     \\ 
            $\cnw(\texttt{archaeologist})$ & \texttt{biologist}  & high        \\
            \bottomrule
        \end{tabular}
    \end{table}
\end{itemize}
\begin{itemize}[leftmargin=2em]
    \item \texttt{Claire} - Similar to \texttt{Bob}, the best negation set for \texttt{Claire} is \{\texttt{Alice}, \texttt{archaeologist}\}. As shown in table below, ``not \texttt{Alice}'' \textbf{moderately} entails \texttt{Claire}.
    
    \begin{table}[!ht]
        \ra{1.2}
        \setlength{\tabcolsep}{18pt}
        \centering
        \begin{tabular}{@{}lll@{}}
            \toprule
            ``not \texttt{Alice}''         & \texttt{Claire}  & Entailment  \\ \hline
            $\cnw(\texttt{Alice})$         & \texttt{Claire}  & medium      \\ 
            \texttt{human}                 & \texttt{human}   & 1 (max)     \\ 
            $\cnw(\texttt{archaeologist})$ & \texttt{pianist} & medium      \\
            \bottomrule
        \end{tabular}
    \end{table}
\end{itemize}
\begin{itemize}[leftmargin=2em]
    \item \texttt{Daisy} - For \texttt{Daisy} the pet dog, the best negation set of ``not \texttt{Alice}'' is \{\texttt{Alice}, \texttt{human}, \texttt{archaeologist}\}. From the table below, ``not \texttt{Alice}'' only \textbf{slightly} entails \texttt{Daisy}.
    
    \begin{table}[!ht]
        \ra{1.2}
        \setlength{\tabcolsep}{18pt}
        \centering
        \begin{tabular}{@{}lll@{}}
            \toprule
            ``not \texttt{Alice}''         & \texttt{Daisy}  & Entailment  \\ \hline
            $\cnw(\texttt{Alice})$         & \texttt{Daisy}  & medium      \\ 
            $\cnw(\texttt{human})$         & \texttt{dog}    & medium      \\ 
            $\cnw(\texttt{archaeologist})$ & \texttt{pet}    & low         \\
            \bottomrule
        \end{tabular}
    \end{table}
\end{itemize}


\newpage
Therefore, in our proposed framework, someone who is not \texttt{Alice} has the highest chance to be (from most to least likely): \texttt{Bob}, \texttt{Claire} and \texttt{Daisy}, which indeed lines up with human intuition. However, the final result of the negation depends on the weights of the negation sets, determined by the context. Hence, if for some reason the negation set $\{\texttt{Alice, human, archaeologist}\}$ has been determined to be the correct interpretation, then ``not \texttt{Alice}'' might be more closely related to \texttt{Daisy} than \texttt{Claire} after all.

\section{Discussion and future work}
\label{sec:future}
The framework to model conversational negation, proposed in this paper, utilises a new mechanism not currently present in DisCoCirc: the external derivation of the weights required for the mixture of different interpretations of a negation. This is motivated by the observation that disambiguation in language relies on an understanding of the world not necessarily present in the text. In the case of the conversational negation of words, this understanding is captured in a single meaning state derived from existing lexical entailment hierarchies, which we called the worldly context (Section~\ref{sec:worldlyContext}). In the case of conversational negation of sentences and evolving meanings, the weights allow for this understanding to be integrated into the circuit. While Section~\ref{section:probabilitiesUsingEntailment} provides an intuition of how the surrounding text can partially inform the correct choice of weights, they should also take into consideration other contexts such as worldly knowledge, intonation, or the environment of the speaker. One goal for future work is to explore these sources of context and find methods to incorporate that information into the weights. Apart from using the weights to inject context, we would like to explore embedding the sources of context directly within the representation of word meanings; for instance, by building upon the work on conceptual space models of cognition in compositional semantics \cite{bolt2019:conceptual}.

The ideas presented in this paper are built on the intuitions gathered from psychology papers and the experiments performed for the conversational negation of words in \cite{rodatz:2021conversationalnegation}. To empirically validate the framework for conversational negation of sentences and evolving meaning, experiments should be devised.
With the basic intuition for deriving the weights from surrounding sentences being solely presented for grammatically identical sentences, further work needs to be done to generalise this process.

In this paper, we modelled the conversational negation of a sentence as the conversational negation of its constituent words.  We focused on single-word constituents, 1) for the purpose of clarity, and 2) because, in our prior work, we proposed and experimentally validated a framework for conversationally negating a single word~\cite{rodatz:2021conversationalnegation}.  However, our model does not forbid constituents made of multiple words.  For instance, in the following example the negated constituent is ``the key to the garage'':
\begin{center}
    \texttt{This is not \underline{the key to the garage}.\quad- \quad This is a toy.}
\end{center}
If we were to broaden the scope of our conversational negation to generate plausible alternatives to a sentence with a different grammatical structure, we would require a mechanism for parsing and negating constituents made of multiple words.  In this view, it could then be possible to model the negation of non-conjunctive composition of concepts such as the compositional explanation of the ``pet fish'' phenomenon by Coecke and Lewis~\cite{coecke:2016petfish}.

Although we have devised a treatment for modelling the conversational negation of a sentence as the conversational negations of its constituent words then composed together grammatically, there are other forms of negation in natural language.
Since Klima~\cite{klima:1964} and Jackendoff~\cite{jackendoff:1969}, most linguists have treated the form of negation we model --- constituent negation --- as distinct from sentential negation~\cite{sep-negation}. Consider the sentence \texttt{I did not walk my dog.} Constituent negation invokes the collective mixture of possible alternative interpretations: that someone else walked the dog, I cuddled the dog, it was my friend's dog, etc. Instead, it may be that we simply want to negate the sentence as a whole --- it is untrue that I walked my dog. Without experiment, it is unclear the extent to which logical negation of the sentence meaning alone suffices to model sentential negation.

Another related challenge is to formalise a mathematical model of the logic underlying conversational NOT, AND, and OR. This requires investigating the extent to which boolean logic holds in a setting known not to follow boolean logic.  A long-term goal would be to extend the conversational negation process to a \emph{conversational logic} process, compatible with compositional distributional semantics, particularly its properties with regards to entailment.

\section*{Acknowledgements}
We would like to thank Vincent Wang for the insightful discussions.
We thank the anonymous reviewers for their helpful feedback.
Lia Yeh gratefully acknowledges funding from the Oxford-Basil Reeve Graduate Scholarship at Oriel College in partnership with the Clarendon Fund.

\bibliographystyle{eptcs}
\bibliography{biblio.bib}

\end{document}